# Multi-Modal Generative Fuzzy System: Fuzzy Inference Guided Large Model Interactive Question Answering Framework

Hailong Yang, Jianqi Wang, Guanjin Wang, *Senior Member*, *IEEE* and Zhaohong Deng, *Senior Member, IEEE*

*Abstract*—**In Multimodal Question Answering (MQA), models are required to jointly encode and integrate heterogeneous information from multiple modalities, including text, images, and speech, to perform complex semantic reasoning and decision making. Despite recent advances, existing approaches, including traditional deep learning models and Large Models (LMs) or prompt-based frameworks, continue to face several critical challenges. First, modality bias arises from discrepancies in feature distributions across different modalities, which limits effective cross modal collaborative understanding. Second, many questions require knowledge drawn from multiple domains, introducing significant uncertainty. Third, current methods often rely on shallow semantic matching, resulting in limited reasoning depth an reduced interpretability. To address these issues, inspired by the traditional fuzzy system (FS) framework, we propose a fuzzy-inference-guided multimodal generative architecture termed the Multi-Modal Generative Fuzzy System (MMGFS). The main contributions of MMGFS are two folds. First, it alleviates modality bias through a multimodal collaborative rumination mechanism. Second, it introduces fuzzy rules and a multi-hop inference mechanism to support cross-domain knowledge fusion and hierarchical reasoning, thereby strengthening uncertainty modelling and deepening semantic understanding. We conduct comprehensive evaluations on open-domain question answering datasets, including MultimodalQA and WebQA, as well as domain-specific benchmarks, including BioMol-VQA and EHRxQA. Experimental results demonstrate that MMGFS consistently outperforms existing methods across multiple datasets. It effectively mitigates modality bias and question uncertainty while achieving superior performance in answer accuracy, consistency, and generalization.**

## I. INTRODUCTION

Multimodal Question Answering (MQA) has emerged as an important research direction in artificial intelligence, aiming to develop effective joint modeling techniques across multiple modalities [1]. It has been widely studied in practical scenarios, including medical image question answering [2], scientific literature retrieval and question answering [3], and Visual Question Answering (VQA) [4], [5].

The objective of MQA is to enable models to jointly understand and process heterogeneous information sources, including text, images, and speech, and to perform cross modal reasoning and answer generation based on integrated understanding. This requires models to extract representative features from each modality while learning meaningful semantic correspondences across modalities to support effective information fusion and multi-step reasoning. Consequently, MQA plays a crucial role in enhancing the comprehensive understanding and decision-making capabilities of artificial intelligence systems.

In MQA tasks, early representative approaches were primarily based on deep learning models. These methods typically employed convolutional neural networks (CNNs) to extract visual features from images and recurrent neural networks (RNNs) to encode the semantic representations of textual questions. The visual and textual features were then fused in a joint space through concatenation, weighted combination, or bilinear pooling [4].With the continuous advancement of deep learning, attention mechanisms were introduced into MQA to improve the selection and alignment of critical cross-modal information [6]. For instance, Stacked Attention Networks (SAN) proposed by Yang *et al.* [7] leverage question-guided attention to focus on image regions most relevant to the textual query, thereby improving answer prediction accuracy. In addition, efficient multimodal interaction methods [8], [9] have further strengthened the integration of visual and linguistic information. Although these approaches achieve promising performance on domain-specific datasets, their robustness and generalization in open-domain question answering tasks remain limited, mainly due to the lack of explicit modeling of complex semantic relationships and logical reasoning mechanisms. With the emergence of the pretraining paradigm, MQA approaches based on cross-modal Transformers and large multimodal models have been extensively studied. These methods are typically pre-trained on large-scale image–text pairs to learn cross-modal alignment and joint semantic representations, followed by task-specific fine-tuning. Contrastive learning techniques [10], [11] further

This work was supported in part by the National Key R&D Program of China under Grant 2022YFE0112400, and in part by the National Natural Science Foundation of China under Grant 62176105. (Corresponding author: Zhaohong Deng).

H. Yang, J. Wang and Z. Deng are with the School of Artificial Intelligence and Computer Science, Jiangnan University, Wuxi 214122, China and Engineering Research Center of Intelligent Technology for Healthcare, Ministry of Education, Wuxi 214122, China. (e-mail: yanghailong@stu.jiangnan.edu.cn; wangjianqi@stu.jiangnan.edu.cn; dengzhaohong@jiangnan.edu.cn).

Guanjin Wang is with the School of Information Technology, Murdoch University, WA, 6150, Australia. (e-mail: Guanjin.Wang@murdoch.edu.au).

enhance image–text alignment in a shared semantic space. More recently, rapid advances in large models (LMs), including Large Language Models (LLMs) and Vision–Language Models (VLMs) have shifted MQA research from conventional feature fusion strategies toward unified LM-centered frameworks. By leveraging cross-modal alignment and the strong reasoning capabilities of LMs, these approaches achieve superior performance in open-ended question answering, multi-turn dialogue, and zero-shot reasoning scenarios [12], [13], [14]. Nevertheless, as the scale and heterogeneity of multimodal inputs increase, existing MQA methods generally lack explicit mechanisms to model the effectiveness and reliability of individual modalities. Consequently, they cannot dynamically regulate modality contributions according to task requirements and contextual variations, limiting their robustness and adaptability in complex open-domain settings. Although existing methods in MQA have achieved notable success from different perspectives, several fundamental challenges remain. These can be summarized as follows.

(1) Aligning heterogeneous features from different modalities (e.g., images, text, and audio) within a shared representation space remains challenging, particularly for fine-grained relationships such as spatial layouts and attribute associations. Existing approaches, largely based on cross-modal attention mechanisms, often fail to capture multi-entity and multi-relation interactions, leading to degraded performance in complex scenarios.

(2) MQA tasks typically span multiple domains, where the importance and relevance of each domain may vary significantly. This inherent multi-domain complexity introduces significant uncertainty that is difficult to address from a single-domain perspective.

(3) Most existing MQA methods rely on shallow feature alignment or pattern matching, which makes them inadequate for multi-hop reasoning and complex logical inference. As a result, they often perform poorly on tasks requiring long-chain reasoning, implicit semantic inference, or external knowledge integration, thereby limiting their generalization and interpretability in real-world applications.

To address these challenges, we propose a novel architecture termed the Multi-Modal Generative Fuzzy System (MMGFS). Inspired by classical fuzzy system architectures, MMGFS can be regarded as a fuzzy-reasoning-guided LM interactive framework. It aligns multimodal inputs into natural language representations and mitigates modality bias through collaborative and rumination mechanisms. Furthermore, MMGFS performs domain fuzzification and rule fusion over questions, leverages fuzzy reasoning to explicitly handle domain-level uncertainty. In addition, MMGFS supports deep semantic reasoning for decision-making via multi-hop fuzzy rule inference. Collectively, these mechanisms enhance both the robustness of MMGFS in MQA tasks and the interpretability of its reasoning process.

The main contributions of this work are summarized as follows.

(1) We propose, for the first time, the concept of a Multi-Modal Generative Fuzzy System (MMGFS). By deeply integrating fuzzy logic with LM-based interactive mechanisms, MMGFS enhances the ability of LMs to perceive and reason under uncertainty. It also extends the design paradigm of large-model-based systems and provides theoretical support for intelligent decision-making in complex and dynamic environments.

(2) From the perspective of fuzzy systems, MMGFS represents an effective extension of classical fuzzy architectures to MQA and related multimodal reasoning tasks, thereby broadening both the theoretical scope and practical applicability.

(3) We introduce a multimodal collaborative rumination mechanism within MMGFS. This mechanism dynamically aligns and iteratively refines multimodal information, including text, images, sequences, and tables at the natural language semantic level, alleviating modality bias and enabling effective cross-modal collaboration and knowledge fusion.

(4) We propose a fuzzy-rule-driven hierarchical knowledge fusion and dynamic reasoning mechanism, which substantially improves MMGFS's capability to handle uncertainty and perform deep semantic reasoning in multi-domain and multimodal tasks.

The remainder of this paper is organized as follows. Section II reviews related work. Section III presents the proposed MMGFS framework, along with detailed analysis and implementation. Section IV reports the experimental setup and results. Finally, Section V concludes the paper and outlines future research directions.

## II. Related Work

### *A. Deep Models for Multimodal Question Answering*

In recent years, MQA has achieved significant progress in interpretable reasoning, knowledge enhancement, and cross-modal fusion. For example, Chen *et al.* proposed MuRAG [15], a retrieval-augmented generative framework that integrates image and text retrieval with generative models, demonstrating strong generalization in open-domain visual question answering (VQA). Khader *et al.* [16] developed a large-scale Transformer-based multimodal model that integrates medical diagnoses, chest imaging, and clinical parameters, highlighting the potential of unified modeling of imaging, textual, and clinical data and achieving improved accuracy and robustness in medical diagnostic QA. Zhou *et al.* proposed Transformer-based cross-modal alignment models, fusion-based VQA models, and knowledge-enhanced QA frameworks [17], further advancing both performance and interpretability of multimodal QA systems. Yu *et al.* introduced Solar [18], a structured knowledge-driven retrieval–generation framework that exhibits promising capability in multi-hop reasoning and complex question answering. Liu *et al.* proposed MMHQA-ICL [19], which enables cross-modal retrieval and evidence aggregation across text, tables, and images through unified multimodal in-context learning. Wang *et al.* presented VQA-GNN [20], which models cross-modal knowledge graphs using graph neural networks for entity- and relation-level reasoning between vision and language. Yang *et al.* proposed SKURG [21], which incorporates an entity-centric fusion encoder to construct multimodal evidence as an input-intrinsic knowledge graph and aligns it into a unified semantic space, thereby alleviating modality bias in cross-modal retrieval.

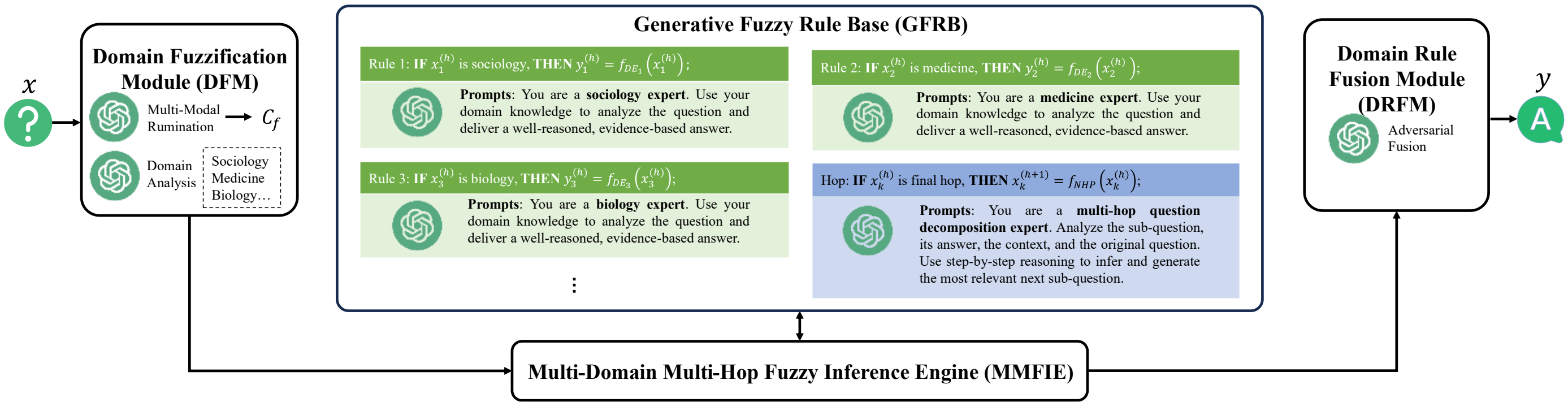

Fig. 1. System framework of MMGFS. The framework comprises the Domain Fuzzification Module (DFM), the Fuzzy Inference Engine (FIE), the Generative Fuzzy Rule Base (GFRB) based on LM interactive components, and the Domain Rule Fusion Module (DRFM). The DFM primarily includes the Multi-Modal Rumination (MMR) submodule and the Domain Analysis (DA) submodule. Within the GFRB rule base, DE is the domain expert component based on LM interactions, and NHP is the Next-Hop Problem handling component. The DRFM module functions to generate results by fusing domain rules based on LM interactions.

Despite these advances, traditional deep QA models remain heavily dependent on task-specific training data, resulting in limited internal knowledge representations and coverage and reduced effectiveness in handling complex questions in open-world scenarios. Moreover, such models face challenges in joint representation learning, cross-modal alignment, and multimodal reasoning, which constrain their applicability in real-world applications.

### *B. Large-Model Interactive Systems for Multimodal Question Answering*

With the rapid advancement of LLMs and VLMs, their applications have expanded across a wide range of tasks. To address increasingly complex problems, researchers have proposed LLM/VLM-based interactive systems [22], [23], [24].

Research in multimodal question answering and reasoning is now shifting from single-model paradigms toward LM-based interactions and controllable reasoning pathways. Early studies primarily focused on the "prompt–action–tool" paradigm and zero-shot generalization. For instance, Li *et al.* proposed AI-VQA [25], which enhances interpretability by introducing interactive mechanisms with VLMs and explicitly presenting intermediate reasoning steps and supporting evidence. Yang *et al.* introduced MM-REACT [26], which realizes coordination between ChatGPT and external tools through prompt-driven perception, planning, and execution chains. Guo *et al.* introduced Img2LLM [27], which improves zero-shot VQA transferability via a ViT-based model and investigates agent specialization and dialectical reasoning mechanisms for VQA from the perspectives of collaboration and information integration. Jiang *et al.* [28] proposed an adaptive multi-LM interactive VQA framework (Multi-Agent VQA), which significantly boosts VQA performance through cooperative interactions among multiple LMs in a zero-shot setting without fine-tuning. Singh *et al.* developed MEQA [29], which securely and efficiently handles predictive queries from heterogeneous data lakes and model repositories by jointly invoking large language models and retrieval-augmented mechanisms. Wu *et al.* [30] proposed a multimodal QA system named RopMura, which employs an intelligent router to select the most suitable knowledge agents and a planner to decompose complex multi-hop questions into manageable steps, thereby enabling efficient and accurate handling of both single-hop and multi-hop cross-domain queries. Shi *et al.* introduced MuMA-ToM [31], which incorporates theory-of-mind modeling into multi-role interactions to enhance uncertainty modeling capabilities. Liu *et al.* proposed a hierarchical multimodal RAG framework, HM-RAG [32], which integrates answers through complex query decomposition, multi-source parallel retrieval (covering text, graph structures, and web data), consistency voting, and expert refinement. Rajput *et al.* presented MAMMQA [33], which decomposes questions, performs cross-modal retrieval and fusion, and finally integrates results using a text-based large language model, achieving superior transparency, accuracy, and robustness compared to existing multimodal QA approaches. In the medical and radiology domains, Zhou *et al.* proposed MAM [34] and related multimodal interactive systems, which enhance multi-step diagnosis and evidence tracing through role-based interactive design and modular collaboration.

Overall, LM-based MQA methods have evolved from simple prompt-based paradigms to interactive systems with planning, collaboration, and tool-use capabilities, enabling stronger multi-step reasoning and cross-modal evidence integration across diverse domains. However, existing approaches still rely heavily on LMs for reasoning and generation, which could lead to unstable reasoning processes. In addition, multimodal fusion, conflict resolution, and the explicit modeling of uncertainty and error propagation remain insufficiently addressed.

### *C. Fuzzy Reasoning and Fuzzy Systems*

Zadeh first introduced the concept of fuzzy sets in 1965 [2], laying the theoretical foundation for the subsequent development of fuzzy theory. Building upon this foundation, rule-based fuzzy systems have been extensively studied and refined. In the supplementary material (Part 1), we provide a detailed description of two representative rule-based fuzzy systems, namely the Takagi–Sugeno–Kang (TSK) fuzzy system [35] and the Mamdani fuzzy system [36].

In the development of question answering methods, rule-based research has consistently played a central role. Traditional rule-based systems (RBS) [37] provide a clear and interpretable reasoning framework for QA tasks; however, they exhibit limited flexibility and adaptability when handling complex tasks. To enhance system adaptability, researchers have introduced fuzzy rules and weighting mechanisms to enhance system adaptability [38]. In fuzzy rule-based classification systems, the incorporation of rule weights can significantly affect overall performance. For example, Navin *et al.* proposed an interpretable medical diagnostic classification method based on fuzzy rules [39], which incorporates weighting schemes and uncertainty handling to improve both model interpretability and diagnostic accuracy.

Regarding the automatic discovery of QA reasoning rules, related studies have explored the integration of rule induction and statistical methods to mitigate the limitations of purely rule-based approaches in open-domain question answering. For instance, Mohammed *et al.* proposed the QArabPro system [40], demonstrating the effectiveness of a rule-based Arabic reading comprehension QA. With the development of LLMs, Zhang *et al.* introduced Rule-KBQA [41], which combines rule-based reasoning with large-scale language models to address complex knowledge-based question answering tasks, thereby highlighting the potential of integrating rule-based methods with statistical learning frameworks.

Despite substantial progress in rule-based methods, fuzzy logic, multimodal modeling, and LLM-driven reasoning, existing approaches still lack a unified framework that jointly integrates multimodal perception, rule-driven reasoning, and collaborative multi-LM interactions. This gap constrains the practical deployment of QA systems in high-demand scenarios such as medical diagnosis, cross-domain knowledge integration, and complex decision-making. Addressing this limitation requires the development of unified and integrated frameworks capable of supporting more robust, interpretable, and generalizable question answering systems.

## III. Multimodal Generative Fuzzy System (MMGFS)

### *A. Framework Overview*

MMGFS explicitly models uncertainty, fuzziness, and conflicting information through fuzzy logic, enabling multi-LM interactive components to collaboratively perceive, reason, and make decisions in dynamic environments, thereby achieving adaptive, robust, and interpretable intelligent behavior. The MMGFS framework consists of four core modules: the Domain Fuzzification Module (DFM), the Generative Fuzzy Rule Base (GFRB), the Multi-Domain Multi-Hop Fuzzy Inference Engine (MMFIE), and the Domain Rule Fusion Module (DRFM), as illustrated in Fig.1.

Within the MMGFS framework, multiple LM-based interactive components are involved, each implemented using an “LLM/VLM + Prompt” paradigm. The prompts are designed based on historical context and generation objectives, and their generalized form expressed in Eq. (1). In the MMGFS system, the historical context of each component accumulates progressively over time, while their objectives remain constant. The generalized form of the processing performed by these LM-based interactive components is shown in Eq. (2).

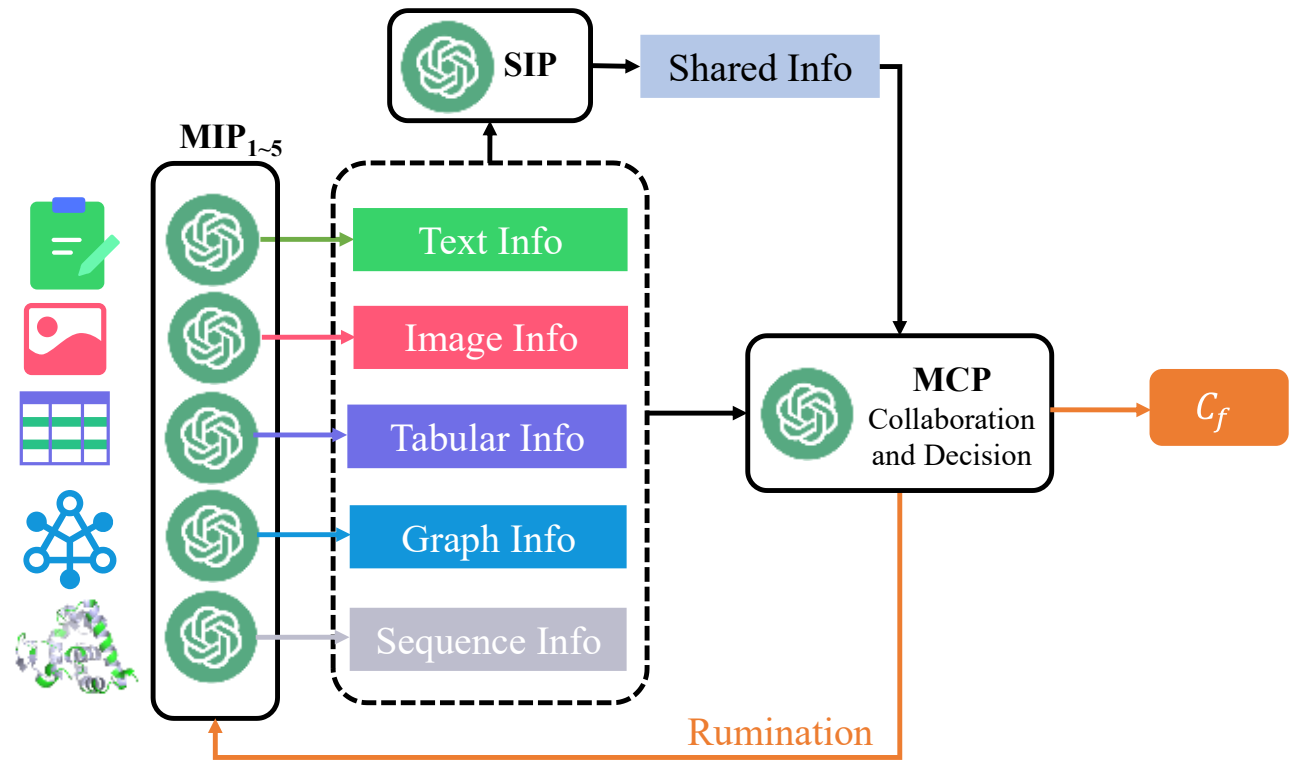

Fig. 2. Multi-Modal Rumination (MMR) submodule. Here, MIP represents the Modal Information Process components for different modalities, SIP is the Shared Information Process component, and MCP is the Modal Collaboration Process component. The orange arrows indicate the rumination mechanism.

$$\mathcal{P} = \psi(C, G) \tag{1}$$

$$A = f_{LM}(I|\mathcal{P}) \tag{2}$$

where $C$ denotes the historical context, $G$ represents the task goal, $\psi$ is the template-based prompt generation function, $\mathcal{P}$ is the generated prompt, and $I$ is the external input, including the user query and data from different modalities related to the query. $f_{LM}$ is the LM interaction-based processing function, and $A$ is the answer generated after LM-based interactions.

The following sections detail the four modules of MMGFS for multimodal question answering.

### *B. Domain Fuzzification Module (DFM)*

To enable efficient analysis and reasoning, MMGFS performs domain-oriented fuzzification of input questions. In the MQA task, besides the question itself, the context contains information from multiple modalities. In the DFM module, we introduce two submodules: Multi-Modal Rumination (MMR) and Domain Analysis (DA). MMR handles multimodal data extraction by identifying shared information and enabling modal collaboration, effectively addressing modality bias. DA, based on LLM interactions, performs domain-specific fuzzification of the input question, addressing the inherent the uncertainty inherent in the problem. The following sections provide a detailed description of the MMR and DA.

#### 1) Multi-Modal Rumination (MMR)

To efficiently handle multimodal data, the DFM module integrates the MMR submodule. MMR simultaneously processes text, images, and tables simultaneously and consists primarily of of three types of components: the Modal Information Process (MIP), the Shared Information Process (SIP), and the Modal Collaboration Process (MCP), as shown in Fig. 2. Specifically, MIP establishes separate components for each modality to handle feature extraction and semantic modeling corresponding to that modality, for example, Text MIP, Image MIP, and Table MIP, thereby ensuring modality-specific optimization and accurate representation of each data type.

In MMR, the modal processing components are not simply combined; instead, they operate through a structured collaboration mechanism to iteratively refine multimodal information. First, the MIP components process data from different modalities, as shown in Eq. (3), and their outputs are subsequently forwarded to the SIP component. The SIP component aggregates and integrates information from all modalities, extracting high-confidence shared information that is consistently supported across multiple modalities, as shown in Eq. (4). Both the original modal information and the extracted shared information are then passed to the MCP component for further processing, as shown in Eq. (5). Guided by the shared information, the MCP component refines the input by removing redundant content within each modality and filtering out low-confidence information that conflicts with the shared information.

$$A_o^m = f_{MIP}^m(I^m|\mathcal{P}_o^m) \tag{3}$$

$$A_s = f_{SIP}(\boldsymbol{A}_M|\mathcal{P}_s) \tag{4}$$

$$C_f = f_{MCP}\big(\boldsymbol{A}_M\big|\psi(\{C_c, A_s\}, G_c)\big) \tag{5}$$

where $I^m$ denotes the original input data of modality $m$, such as text, images, or tables. $A_o^m$ represents the output of the MIP component for modality $m$, where $m = [1, \dots, M]$ and $M$ is the number of modalities supported by the MCP component. $\boldsymbol{A}_M = \{A_o^1, \dots, A_o^m, \dots, A_o^M\}$ denotes the set of outputs from all MIP components. $\mathcal{P}_o^m$ is the prompt used by the MIP for modality $m$, and $f_{MIA}^m(.)$ is the processing function of the MIP component for modality $m$. $A_s$ is the shared information extracted by SIP from the $M$ modalities, $\mathcal{P}_s$ is the prompt for the SIP component, and $f_{SIP}(.)$ is the SIP processing function. $C_f$ denotes the final output of MCP, $f_{MCP}(.)$ is the MCP processing function, $C_c$ is the historical context of the MCP component, and $G_c$ is the generation goal of MCP.

In multimodal data, information often exhibits asymmetry. For example, an image may contain rich semantic content, while the corresponding textual description may be brief or incomplete, leading to an imbalance of information across modalities. To address this issue, we introduce a Rumination Mechanism (RM) in MMR. RM feeds the information $A_c$ processed by MCP back to each MIP as part of the prompt, where it is combined with the original modal data for further refinement and analysis. This allows for deeper discovery of hidden patterns and improves the consistency and completeness of the information. Under the rumination mechanism, Eq. (3) is extended to Eq. (6).

$$A^{m,(n)} = f_{MIP}^m\left(I^m\middle|\psi\left(\left\{C_o^{m,(n)}, C_f^{(n-1)}\right\}, G_o^m\right)\right) \tag{6}$$

where $A_c^{(n-1)}$ denotes the information output by MCP during the $(n-1)$-th rumination, with $n \in \mathbb{N}^+$. $A^{m,(n)}$ is the information output by the modality-specific MIP component for modality $m$ during the $n$-th rumination, performing deep exploration of the input data $I^m$ under the guidance of the shared information. $C_o^{m,(n)}$ represents the historical context at the $n$-th rumination, which grows as the number of rumination iterations increases. $G_o^m$ is the task goal content for the LM interaction, which remains constant throughout.

MCP outputs the final processed result only after verifying that the input information is free from contradictions and semantic errors, thereby achieving high-confidence cross-modal information integration. The algorithmic procedure of the RM mechanism is presented in Algorithm I.

**Algorithm I** Rumination Mechanism
**Inputs**: Modality Data $I^m, m = [1 \dots M], M \in \mathbb{N}^+$
**Output**: Final Context $C_f$

1: Initialization rumination $n \leftarrow 0$
2: $C_f, C_f^{(n)} \in \emptyset$
3: **repeat**
4: **for** *m in* M **do**
5: update modality information $A_o^{m,(n)}$ based on (3)
6: **end for**
7: update shared information $A_s^{(n)}$ based on (4)
8: calculate $C_f^{(n)}$ based on (5)
9: $C_f \leftarrow C_f^{(n)}$
10: $n \leftarrow n + 1$
11: **until** no contradiction
12: **return** $A_c$

### 2) Domain Analysis (DA)

In MQA tasks, user questions often involve domain-related uncertainties as well as logically complex issues. In this study, we address the former using fuzzy reasoning and handle the latter through multi-hop decomposition. After obtaining sufficient background information about the question, we perform domain-specific fuzzification of the input, which mainly involves two steps: (1) domain identification, and (2) generating the first-hop question for each domain, as illustrated in Fig. 3.

Based on the above approach, we construct the DA submodule with two LLM interactive components: (1) a component for analyzing the domains involved in a question based on the multimodal context, as shown in Eq. (7); and (2) a component for generating the first-hop question for each domain, as shown in Eq. (8).

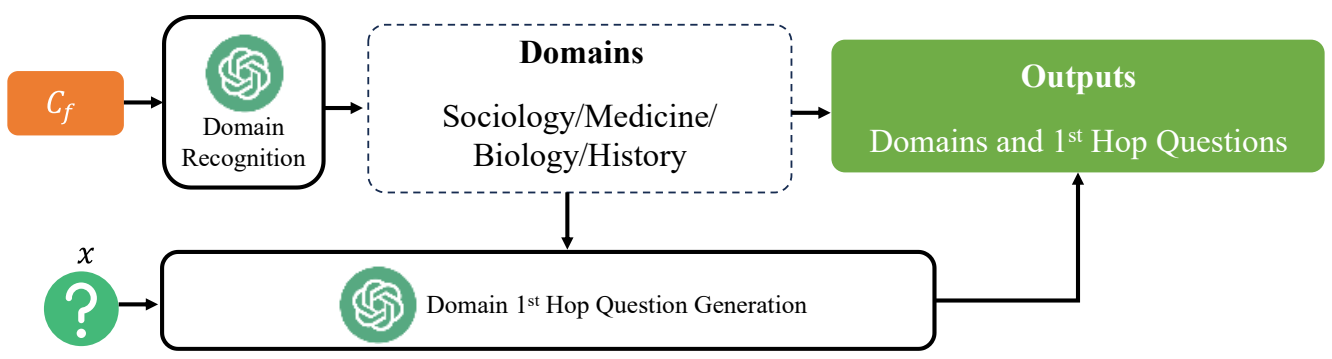

Fig. 3. Structure of the Domain Analysis (DA) submodule. DA mainly consists of the Domain Identification component and the First-Hop question Generation component.

$$\boldsymbol{d} = f_{DA}^{(1)}\left(x\middle|\psi\left(C_f, G_d\right)\right) \tag{7}$$

$$\boldsymbol{x}^{(1)} = f_{DA}^{(2)}\left(x\middle|\psi\left(C_f, \boldsymbol{d}, G_h\right)\right) \tag{8}$$

where $x$ is the user's original input question. $f_{DA}^{(1)}(.)$ is the processing function of the DA submodule for analyzing the domains of the question, and $f_{DA}^{(2)}(.)$ is the processing function for generating the first-hop question. $C_f$ represents the background information processed by the MMR submodule. $G_d$ is the objective of the domain analysis, and $G_h$ is the task goal for generating the first-hop question. $\boldsymbol{d}$ denotes the knowledge domains involved in question $x$, e.g., $\boldsymbol{d} = [d_1, \dots, d_K]$. $\boldsymbol{x}^{(1)}$ is the set of first-hop questions for each domain, $\boldsymbol{x}^{(1)} = [x_1^{(1)}, \dots, x_K^{(1)}]$, where $K$ is the total number of domains involved in $x$, with $K \in \mathbb{N}^+$.

*C. Generative Fuzzy Rule Base (GFRB)*

Traditional fuzzy systems employ rule bases whose consequents are typically uncertainty-handling functions, such as the linear functions used in TSK-type fuzzy systems. While these formulations are relatively easy to learn and analyze, their expressiveness of the encoded expert knowledge remains limited, making them insufficient for complex multimodal question-answering tasks. To overcome this limitation, we introduce LLM-driven interactive rules and propose a novel form of generative fuzzy rule. Similar to classical fuzzy rules, a generative fuzzy rule consists of an antecedent and a consequent; however, its antecedent computes membership degrees through LLM-based semantic reasoning, while its consequent is constructed via LLM-driven domain expert interactions, enabling it to directly generate answers or support next-hop reasoning in MQA tasks.

The set of generative fuzzy rules forms the Generative Fuzzy Rule Base (GFRB). This rule base contains $K$ domain rules, each designed to address specific problems within a particular domain and provide corresponding solutions. In addition, the GFRB includes a Hop Rule, which is used to generate the next-hop questions. The formal structure of the rules is shown in Eq. (9). Specifically, the antecedent of each domain rule leverages LLM semantic reasoning to determine the membership degree between the input question and the domain represented by the rule, while the consequent generates the corresponding answer to the current-hop question through the DE component. The rule base contains only one Hop Rule, whose antecedent determines whether the current hop is the last, and whose consequent is responsible for generating the next-hop questions for all domain rules.

$$\begin{aligned} &Rule\ 1\text{: } \mathbf{IF}\ x_1^{(h)}\ is\ sociology\text{, } \mathbf{THEN}\ y_1^{(h)} \\ &= f_{DE_1}\left(x_1^{(h)} \middle| \psi\left(\left\{C_f, y_1^{(h-1)}\right\}, G_{d_1}\right)\right); \\ &\dots \\ &Rule\ K\text{: } \mathbf{IF}\ x_K^{(h)}\ is\ history\text{, } \mathbf{THEN}\ y_K^{(h)} \\ &= f_{DE_K}\left(x_K^{(h)} \middle| \psi\left(\left\{C_f, y_K^{(h-1)}\right\}, G_{d_K}\right)\right); \\ &Hop\text{: } \mathbf{IF}\ x_k^{(h)}\ is\ final\ hop\text{, } \mathbf{THEN}\ x_k^{(h+1)} \\ &= f_{NHP}\left(x_k^{(h)} \middle| \psi\left(\left\{C_f, y_k^{(h)}, x\right\}, G_h\right)\right); \end{aligned} \tag{9}$$

where, Rule $k$ refers to the $k$-th domain rule, with $k = [1, \dots, K]$. $h$ denotes the hop number in the multi-hop questions. $y_k^{(h-1)}$ is the answer to the previous-hop question in domain $k$, and $y_k^{(h)}$ is the answer to the current-hop question in domain $k$. $x_k^{(h)}$ is the current-hop question for domain $k$, and $x_k^{(h+1)}$ is the next-hop question for domain $k$. $G_{d_k}$ represents the objective of the domain rule, and $G_h$ is the goal for generating the first-hop question.

*D. Multi-Domain Multi-Hop Fuzzy Inference Engine (MMFIE)*

The Multi-Domain Multi-Hop Fuzzy Inference Engine (MMFIE) is an LLM interactive rule inference module that focuses on semantic-level reasoning as shown in Fig.4. Its inference process primarily consists of domain rule reasoning and Hop rule reasoning. Domain rule reasoning aims to provide answers to the same question across multiple domains, helping to mitigate uncertainties arising from domain differences. Hop rule reasoning, on the other hand, focuses on contextual reasoning: by determining whether the current-hop question requires the generation of a next-hop question, it incrementally constructs a multi-hop reasoning process based on rules, continuously uncovering deeper semantic information. The following sections provide a detailed explanation of domain rule reasoning and Hop rule reasoning.

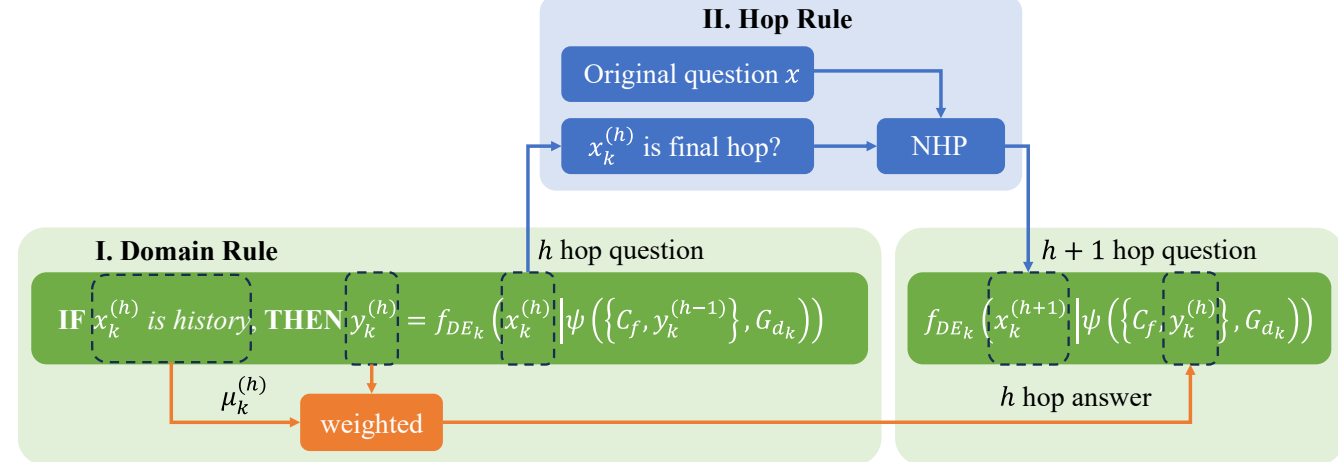

Fig. 4. Inference process of the MMFIE fuzzy inference engine. It consists of two key steps: I. Domain Rule–Based Reasoning, which outputs answers for the next-hop questions; II. Hop Rule–Based Reasoning, which generates the next-hop questions.

**1) Domain Rule Reasoning**

In traditional fuzzy inference systems, membership degrees in the antecedents are calculated based on feature distances or similarity measures, which are insufficient to capture semantic-level membership relationships. Modern LLMs, such as the OpenAI GPT series, are general-purpose LMs with rich domain knowledge and strong semantic reasoning capabilities, enabling membership computation across multiple domains. Therefore, in MMFIE, we use LLM-based semantic reasoning to estimate the domain membership of input questions, as shown in Eq. (10). The resulting membership is expressed in natural language and is referred to as domain membership terminology.

In MMFIE, each hop question is answered by the DE component in the rule consequent. The output is weighted according to its corresponding domain membership before being passed to the domain rules of the next hop, as shown in Eq. (11). At the semantic level, this weighting operation is realized by annotating the generated answer, that is, prepending a short phrase that reflects the degree of domain membership. Its generalized form is expressed in Eq. (12).

$$\mu_k^{(h)} = f_\mu\left(x_k^{(h)} \middle| \psi(d_k, G_\mu)\right) \tag{10}$$

$$y_k^{(h)} = f_{DE_k}\left(x_k^{(h)} \middle| \psi\left(\left\{C_f, y_k^{(h-1)}\right\}, G_{d_k}\right)\right) \tag{11}$$

$$y_k^{(h)} \leftarrow Concat\left(\mu_k^{(h)}, y_k^{(h)}\right) \tag{12}$$

where $d_k$ denotes the $k$-th domain, such as {*sociology, medicine, biology, …*}. $G_\mu$ represents the objective for membership calculation, and $f_\mu(.)$ denotes the LLM-based membership computation function. $\mu_k^{(h)}$ represents the domain membership of the question, with the terminology set {*High (H), Upper-Medium (UM), Medium (M), Lower-Medium (LM), Low (L), Very-Low (VL)*}, $Concat(.)$ denotes the concatenation function.

**2) Hop Rule Reasoning**

The antecedent of the Hop Rule is used to calculate the membership degree of the current input question with respect to whether it corresponds to the last-hop question, as shown in Eq. (13). Its computation is similar to that of domain membership, as it relies on LLM-based semantic reasoning. When this membership degree is classified "High," it indicates that the current-hop question is highly likely to be the last hop, and no additional hop questions are generated. For other membership levels, the subsequent hop question is generated to continue the multi-hop reasoning process, as shown in Eq. (14).

$$\mu_f = f_\mu\left(x_k^{(h)} \middle| \psi\left(\left\{C_f, y_k^{(h)}, x\right\}, G_f\right)\right) \tag{13}$$

$$x_k^{(h+1)} = \begin{cases} \emptyset & , \mu_f = H \\ f_{NHP}\left(x_k^{(h)} \middle| \psi\left(\left\{C_f, y_k^{(h)}, x\right\}, G_h\right)\right) & , Otherwise \end{cases} \tag{14}$$

where $\mathcal{P}_h$ is the prompt for the NHP component, $\mu_f$ denotes the membership degree of the input current-hop question being the last hop, $G_f$ is the task objective for calculating the last-hop membership, and $G_h$ is the task objective for generating the next-hop question.

In MMFIE, domain rule reasoning and Hop rule reasoning operate in a complementary manner. MMFIE first performs reasoning based on multiple domain rules and then generates the questions required for the next hop through the Hop rule. It should be noted that domain rules applied in the next hop are structurally identical to those in the current hop, although their input and output content differ. All domain rules are governed by a single Hop rule. The detailed procedure of the MMFIE inference algorithm is presented in Algorithm II.

**Algorithm II** Multi-Modal Fuzzy Rule Inference

**Inputs**: original question $x$, domains $\boldsymbol{d}$, 1st hop questions $\boldsymbol{x}^{(1)}$
**Output**: Final answers $\boldsymbol{y}$

1: $K \in \mathbb{N}^+$ // total domains
2: $h \in \mathbb{N}^+$ // hop number
3: $\boldsymbol{d} = [d_1 \dots d_K]$
4: $\boldsymbol{x}^{(1)} = [x_1^{(1)} \dots x_K^{(1)}]$
5: **for** $k$ in $K$ **do**
6: $h \leftarrow 1$
7: **repeat**
8: calculate $\mu_k^{(h)}$ based on (9)
9: gain the answer $y_k^{(h)}$ of the $k$-hop question $x_k^{(h)}$ using DE
10: weight the answer on (10)
11: **if** $x_k^{(h)}$ is final hop **do**
12: $x_k^{(h+1)} = \emptyset$
13: **else**
14: gain the next-hop question $x_k^{(h+1)}$ using NHP
15: $h \leftarrow h + 1$
16: **end if**
17: **until** final hop
18: $y_k = y_k^{(h)}$ // domain answer
19: **end for**
20: $\boldsymbol{y} = [y_1 \dots y_K]$
21: **return** $\boldsymbol{y}$

### E. Domain Rule Fusion Module (DRFM)

The Domain Rule Fusion Module (DRFM) is responsible for integrating the answers generated by multiple domain rules into a final result as shown in Fig. 5. Considering that LLMs may produce hallucinations or introduce semantic errors, DRFM incorporates a semantic adversarial mechanism during the fusion process to enhance reliability and robustness.

The adversarial fusion process consists of two steps: (1) voting among the candidate answers to retain the one with receiving the highest votes, as shown in Eq. (15); and (2) filtering out answers with low domain relevance based on their corresponding domain membership values, as shown in Eq. (16). Through this process, high-confidence answers are selected and subsequently combined to form the final answer. It is worth noting that if there is only one candidate answer with the highest votes, it is directly adopted as the final output. When multiple candidates tie for the highest votes, further filtering is performed by comparing their membership degrees. If multiple candidates still remain after this step, they are fused through LLM interactions to generate the consolidated final output.

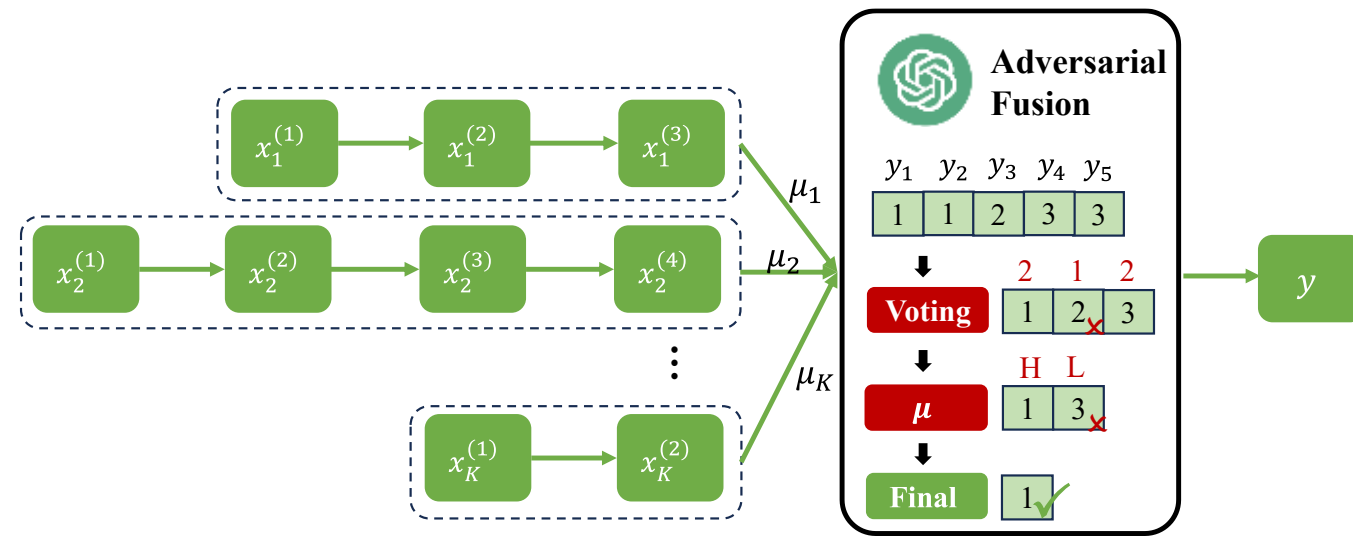

Fig. 5. Domain Rule Fusion Module: Each domain rule produces a domain-specific answer after multiple hops. The $K$ domain answers are then fused to generate the final answer $y$.

$$\hat{\boldsymbol{y}} = \underset{y \epsilon \boldsymbol{y}}{arg\,max} \sum_{i=1}^{K} \mathbb{I}(y_i = y) \tag{15}$$

$$y_f = \underset{y \epsilon \hat{\boldsymbol{y}}}{arg\,max} \max_{i \in \mathcal{I}(\hat{\boldsymbol{y}})} \mu_i \tag{16}$$

where $\mathbb{I}(.)$ is the voting indicator function, $\mathcal{I}(.)$ is the answer indexing function, and $y_f$ denotes the final answer after adversarial fusion.

## IV. EXPERIMENTS

### A. Experimental Setup

#### 1) Datasets

The datasets used in the experiments are primarily divided into two categories: open-domain question-answering datasets and domain-specific question-answering datasets. As shown in Fig. 6, MultimodalQA [42] and WebQA [43] belong to the open-domain QA category, while BioMol-MQA [44] and EHRxQA [45] are domain-specific QA datasets. The figure also illustrates the sample distributions for each dataset. Table S1 in

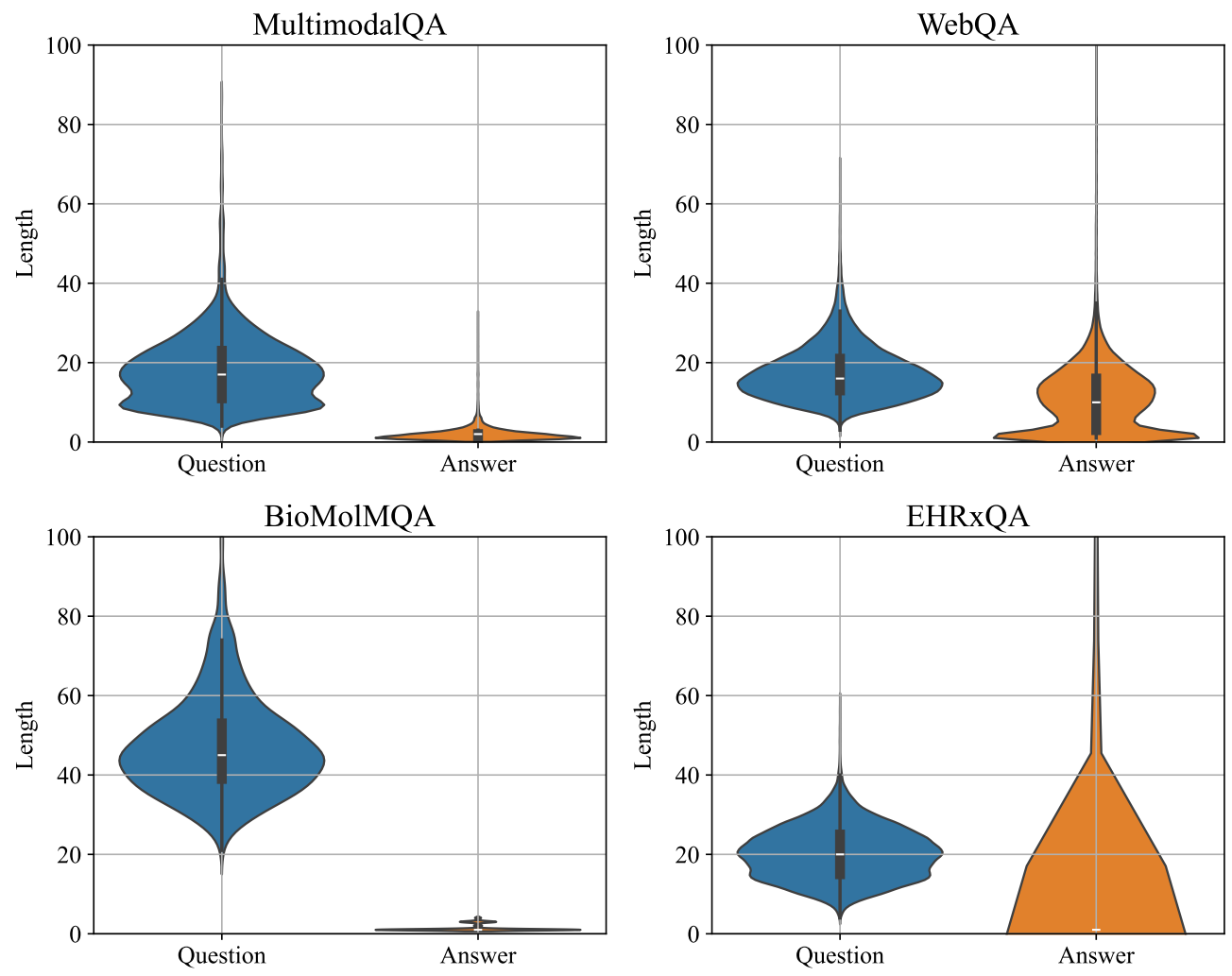

Fig. 6. Length distributions of questions and answers in the datasets used for the experiments.

TABLE I
PERFORMANCE COMPARISON ON THE OPEN-DOMAIN QUESTION ANSWERING DATASETS MULTIMODALQA AND WEBQA

| Method | MultiModalQA | | | | | | WebQA | | | | | |
|---|---|---|---|---|---|---|---|---|---|---|---|---|
| | Acc | Recall | F1 | EM | MRR | HR@5 | Acc | Recall | F1 | FL | BLEU | METEOR |
| MuRAG | 55.32 | 60.11 | 57.62 | 51.40 | 30.25 | 40.22 | 47.80 | 49.70 | 48.73 | 50.70 | 49.86 | 45.46 |
| Solar | 62.54 | 69.56 | 65.86 | 59.80 | 34.98 | 45.69 | 59.40 | 60.90 | 60.14 | 58.90 | 50.11 | 46.31 |
| VQA-GNN | 64.22 | 72.55 | 68.13 | 60.25 | 35.99 | 50.22 | 58.66 | 61.58 | 60.08 | 52.66 | 55.66 | 48.75 |
| SKURG | 65.22 | 70.14 | 67.59 | 59.82 | 35.21 | 49.32 | 57.10 | 64.23 | 60.46 | 55.40 | 56.11 | 49.44 |
| MM-REACT | 66.21 | 72.25 | 69.10 | 61.25 | 36.25 | 50.26 | 59.02 | 61.11 | 60.05 | 59.67 | 60.23 | 51.42 |
| Multi-AgentVQA | 71.65 | 72.56 | 72.10 | 63.89 | 38.52 | 51.09 | 60.05 | 62.54 | 61.27 | 62.22 | 62.75 | 53.88 |
| Rule-KBQA | 71.00 | 72.61 | 71.80 | 65.37 | 39.02 | 52.06 | 58.61 | 63.25 | 60.84 | 61.23 | 64.23 | 55.44 |
| HM-RAG | 72.89 | 75.69 | 74.26 | 65.88 | 42.36 | 52.11 | 60.11 | 63.55 | 61.78 | 62.77 | 68.31 | 56.31 |
| MAMMQA | 74.22 | 75.68 | 74.94 | 67.56 | **44.51** | 54.93 | 61.87 | 63.08 | 62.47 | 63.21 | **71.09** | 58.62 |
| MAS-CR | 74.88 | 69.68 | 72.19 | 64.23 | 41.06 | 52.24 | 60.24 | 63.58 | 61.86 | 62.01 | 67.62 | 55.75 |
| **MMGFS** | **75.45** | **79.83** | **77.58** | **69.61** | **44.51** | **55.91** | **62.03** | **67.01** | **64.42** | **65.44** | 71.07 | **59.68** |

TABLE II
PERFORMANCE COMPARISON ON THE DOMAIN-SPECIFIC QUESTION ANSWERING DATASETS BIOMOL-MQA AND EHRXQA

| Method | BioMol-MQA | | | | | | EHRXQA | | | | | |
|---|---|---|---|---|---|---|---|---|---|---|---|---|
| | Acc | Recall | F1 | EM | MRR | HR@5 | Acc | Recall | F1 | FL | BLEU | METEOR |
| MuRAG | 36.98 | 42.30 | 39.46 | 30.12 | 43.51 | 54.01 | 54.26 | 55.23 | 54.74 | 70.25 | 71.29 | 62.46 |
| Solar | 42.22 | 44.55 | 43.35 | 33.69 | 45.36 | 54.24 | 58.31 | 59.36 | 58.83 | 72.58 | 72.30 | 62.62 |
| VQA-GNN | 47.02 | 48.80 | 47.89 | 36.24 | 49.65 | 57.51 | 58.01 | 59.00 | 58.50 | 73.01 | 75.48 | 66.16 |
| SKURG | 49.36 | 50.12 | 49.74 | 39.01 | 52.34 | 60.21 | 58.36 | 57.99 | 58.17 | 75.36 | 78.88 | 68.75 |
| MM-REACT | 50.23 | 51.14 | 50.68 | 39.57 | 59.56 | 65.32 | 59.26 | 60.23 | 59.74 | 78.23 | 78.29 | 70.46 |
| Multi-AgentVQA | 50.64 | 52.00 | 51.31 | 40.02 | 60.35 | 66.21 | 60.25 | 61.85 | 61.04 | 80.04 | 80.32 | 71.56 |
| Rule-KBQA | 52.47 | 56.09 | 54.22 | 41.41 | 61.23 | 67.25 | 62.45 | 63.59 | 63.01 | 84.05 | 79.16 | 72.45 |
| HM-RAG | 53.99 | 57.36 | 55.62 | 42.09 | 63.25 | 68.01 | 63.58 | 65.29 | 64.42 | 85.22 | 78.89 | 73.00 |
| MAMMQA | 55.88 | **60.00** | **57.87** | 43.66 | 64.58 | 69.21 | 65.09 | **72.65** | 68.66 | 87.24 | 80.89 | 76.40 |
| MAS-CR | 53.54 | 57.21 | 55.31 | 42.27 | 62.22 | 68.24 | 63.66 | 68.25 | 65.88 | 89.87 | 78.51 | 74.29 |
| **MMGFS** | **56.19** | 58.55 | 57.35 | **44.58** | **65.01** | **70.61** | **69.21** | 72.57 | **70.85** | **90.54** | **81.03** | **76.46** |

TABLE III
MODULE ABLATION RESULTS ON THE OPEN-DOMAIN QUESTION ANSWERING DATASETS MULTIMODALQA AND WEBQA

| Method | MultimodalQA | | | | | | WebQA | | | | | |
|---|---|---|---|---|---|---|---|---|---|---|---|---|
| | Acc | Recall | F1 | EM | MRR | HR@5 | Acc | Recall | F1 | FL | BLEU | METEOR |
| B | 62.53 | 63.57 | 63.05 | 58.13 | 30.56 | 42.69 | 50.68 | 52.11 | 51.39 | 54.66 | 56.30 | 49.22 |
| B+S | 70.25 | 72.46 | 71.34 | 64.54 | 37.45 | 48.39 | 56.31 | 59.36 | 57.79 | 55.03 | 57.18 | 51.23 |
| B+S+R | 72.23 | 75.89 | 74.01 | 66.19 | 41.88 | 51.04 | 59.68 | 64.97 | 62.21 | 60.55 | 64.24 | 55.41 |
| B+S+R+F | **75.45** | **79.83** | **77.58** | **69.61** | **44.51** | **55.91** | **62.03** | **67.01** | **64.42** | **65.44** | **71.07** | **59.68** |

TABLE IV
MODULE ABLATION RESULTS ON THE DOMAIN-SPECIFIC QUESTION ANSWERING DATASETS EHRXQA AND BIOMOL-MQA

| Method | BioMol-MQA | | | | | | EHRXQA | | | | | |
|---|---|---|---|---|---|---|---|---|---|---|---|---|
| | Acc | Recall | F1 | EM | MRR | HR@5 | Acc | Recall | F1 | FL | BLEU | METEOR |
| B | 49.56 | 50.22 | 49.89 | 31.64 | 50.13 | 56.45 | 55.61 | 54.68 | 55.14 | 50.21 | 55.64 | 49.52 |
| B+S | 53.03 | 54.44 | 53.73 | 37.88 | 57.21 | 62.80 | 62.56 | 63.14 | 62.85 | 52.89 | 59.56 | 52.49 |
| B+S+R | 54.33 | 56.34 | 55.32 | 41.19 | 61.91 | 66.39 | 64.78 | 68.33 | 66.51 | 72.65 | 70.55 | 65.29 |
| B+S+R+F | **56.19** | **58.55** | **57.35** | **44.58** | **65.01** | **70.61** | **69.21** | **72.57** | **70.85** | **90.54** | **81.03** | **76.46** |

Part 2 of the supplementary materials summarize the key statistics of the four datasets, including the total number of samples, involved modality types, question lengths, and answer lengths, and also provides detailed descriptions and download links.

**2) Experimental Settings**

To comprehensively evaluate the proposed MMGFS method, we selected two categories of baseline methods: traditional deep learning methods and LM-interactive methods. The deep learning methods include MuRAG, Solar, VQA-GNN, and SKURG, while the LM-interactive systems include MM-REACT, Multi-Agent VQA, Rule-KBQA, HM-RAG, MAMMQA, and MAS-CR.

In the experiments, to handle different modalities, we selected multiple LLMs and VLMs. For extracting semantics from text, tables, and biological sequences, we used GLM 4.5 (glm-4.5-flash), DeepSeek 3.1 (deepseek-3.1), and GPT-4. To process visual semantics from image modalities, we employed the VLM glm-4v-flash. For medical images, we used VLM-Med (LLaVA-Med-v1.5-Mistral-7B) to extract semantics from CT scans.

The experimental platform is configured as follows. The software environment included Python 3.8.5, Torch 2.3.0, and CUDA 12.1, while the hardware platform consisted of 10 NVIDIA RTX A6000 GPUs, 2 Intel 8352V CPUs, and 256 GB of memory.

We adopt three categories comprising nine evaluation metrics in total. Result consistency is assessed using EM (Exact Match), MRR (Mean Reciprocal Rank), and HR (Hit Rate); result accuracy is evaluated with Acc (Accuracy), Recall, and F1; and result fluency is measured by FL (Fluency), BLEU (Bilingual Evaluation Understudy), and METEOR (Metric for Evaluation of Translation with Explicit ORdering). The detailed computation methods of these evaluation metrics are provided in Part 3 of the supplementary material.

### *B. Performance Comparison*

To comprehensively evaluate the performance of the proposed method, we conducted comparative experiments on open-domain question answering datasets, MultiModalQA and WebQA, as well as domain-specific QA datasets, BioMol-MQA (drug–target interactions) and EHRxQA (electronic health records). Notably, the QA results on all datasets were consistently evaluated using answer accuracy metrics, including Acc, Recall, and F1.

However, different metrics were adopted for evaluating answer matching and answer fluency across datasets. This discrepancy primarily arises from differences in answer length distributions. Specifically, answers in MultiModalQA and BioMol-MQA are predominantly short, typically consisting of one to two tokens. Such short texts are insufficient to effectively reflect linguistic fluency, making answer matching metrics more appropriate for evaluation, such as EM, MRR, and HR@5. In contrast, answers in WebQA and EHRxQA are generally longer and exhibit complete natural language structures, which are better suited for fluency-oriented evaluation metrics, including FL, BLEU, and METEOR.

Tables I and II present the performance comparisons on the open-domain QA datasets and the domain-specific QA datasets, respectively. As shown in Table I, in open-domain QA tasks, MMGFS outperforms the compared methods on the majority of evaluation metrics. Specifically, on the MultiModalQA dataset, MMGFS achieves superior performance over all traditional deep learning methods and LM-interactive approaches in terms of both answer accuracy (Acc) and answer matching metrics. Compared with the best-performing LM-interactive method, MAMMQA, MMGFS improves Acc by 1.23 percentage points, F1 by 2.64, EM by 2.05, and HR@5 by 0.98. On the WebQA dataset, MMGFS achieves a 2.23 improvement in FL over MAMMQA, indicating superior answer fluency compared with all baseline methods. Although MMGFS is slightly lower by 0.02 in BLEU, it outperforms MAMMQA by 1.06 on the BLEU-extended metric METEOR. As shown in Table II, MMGFS consistently outperforms all baseline methods on answer matching–related metrics across both the BioMol-MQA dataset in the bioinformatics domain and the EHRxQA dataset in the medical domain. In terms of answer accuracy, MMGFS performs slightly worse than MAMMQA; for example, its Recall is lower by 1.45 on BioMol-MQA and by 0.08 on EHRxQA. This performance gap is mainly attributed to the fact that MMGFS emphasizes semantic consistency during multimodal feature fusion, while its coverage of long-tail answers is relatively weaker than that of MAMMQA. Nevertheless, on the EHRxQA dataset, MMGFS significantly outperforms MAS-CR, a recent state-of-the-art multi-LM interactive method in the medical domain, achieving an Acc improvement of 5.55.

Overall, MMGFS demonstrates consistently superior performance compared with all LM-interactive methods, while LM-interactive approaches generally outperform traditional methods. These results indicate that, by leveraging multimodal collaboration and a fuzzy rule–based multi-level uncertainty reasoning mechanism, MMGFS exhibits notable advantages in terms of answer accuracy, matching quality, and fluency.

## C. Ablation Study

To analyze the effectiveness of each module and mechanism, we conducted an ablation study on MMGFS. Specifically, we compared the baseline method with its progressively enhanced variants. The baseline refers to directly performing question answering using a LLM after extracting information from each modality, denoted as “B”. “ +S” indicates that shared information extracted from different modalities is incorporated and jointly provided to the LLM as part of the contextual input along with other modal information. “ +R” denotes the introduction of a deep mining and rumination mechanism for multimodal data. “ +F” represents the addition of a fuzzy reasoning mechanism, which mainly includes domain expert components from domain rules and multi-hop reasoning. Tables III and IV present the ablation results of MMGFS on open-domain and domain-specific question answering tasks, respectively.

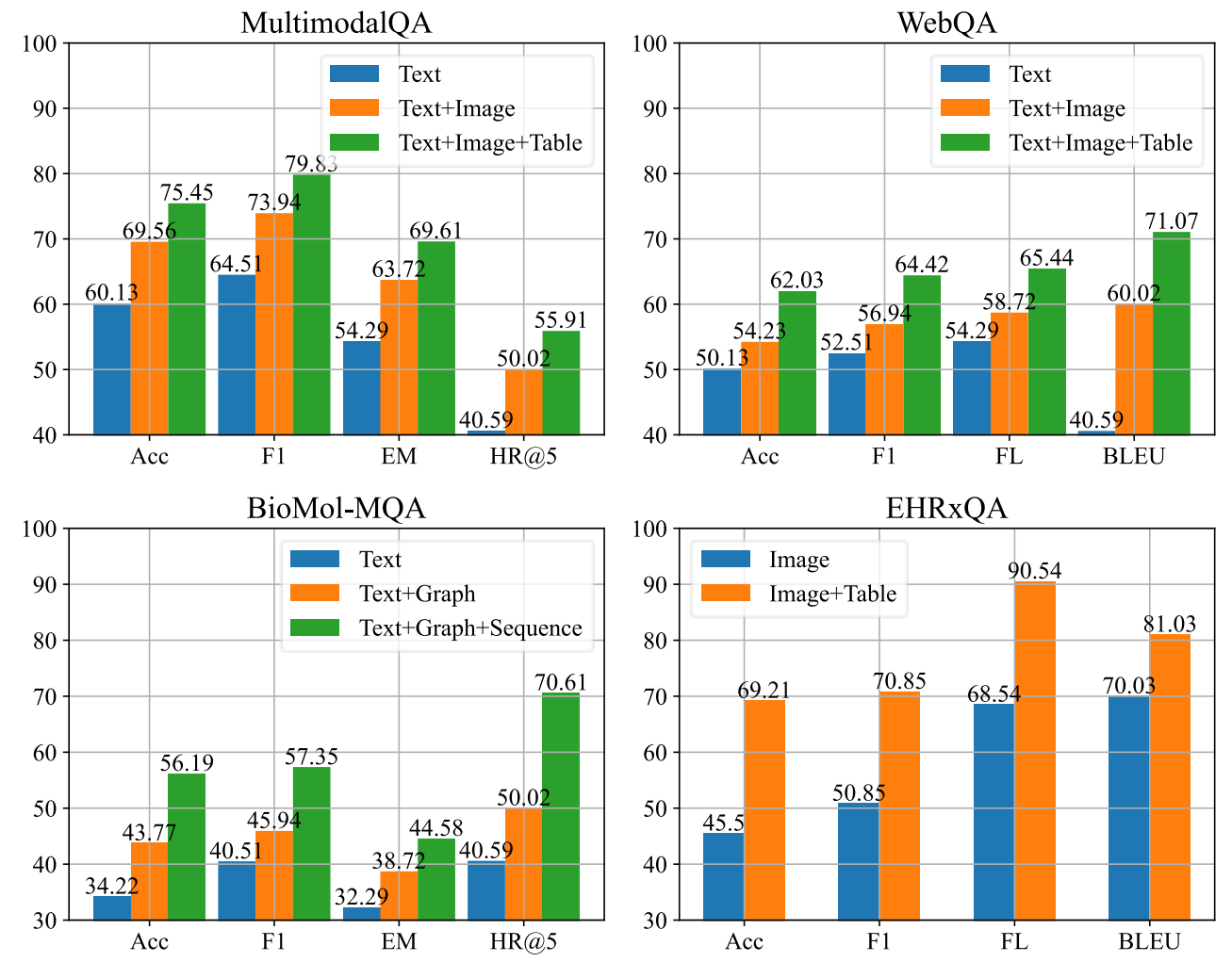

Fig. 7. Multimodal effectiveness analysis. In BioMol-MQA, the Graph modality corresponds to the drug–target interactive graph, and the Sequence modality consists of molecular SMILES sequences and protein sequences. In EHRxQA, the Image modality corresponds to CT images.

As shown in Table III, on the open-domain QA task MultiModalQA, the F1 score demonstrates a steady performance improvement with the incremental introduction of each module. Specifically, incorporating multimodal shared information improves performance by 8.29, further introducing the rumination mechanism for multimodal data yields an additional gain of 2.68, and finally, the inclusion of fuzzy rule–based reasoning results in a further improvement of 3.56. On the WebQA task, a similar trend is observed for the FL metric: adding multimodal shared information improves performance by 0.37, introducing the rumination mechanism leads to a substantial gain of 5.52, and further incorporating fuzzy rule reasoning yields an additional improvement of 4.89.

As shown in Table IV, on the domain-specific BioMol-MQA dataset, adding shared information improves the EM metric by 19.72% over the baseline method B; further introducing the rumination mechanism results in an additional improvement of 8.74%; and incorporating rule-based reasoning on this basis leads to a further gain of 8.23%. A similar pattern is observed on the EHRxQA dataset, where the BLEU score improves by 7.05%, 18.45%, and 14.85%, respectively.

Overall, these results indicate that the shared information, rumination mechanism, and fuzzy rule–based reasoning in MMGFS each make distinct and positive contributions to the final performance. Moreover, these components exhibit strong complementary and synergistic effects: shared information provides the foundation for cross-modal semantic fusion, the rumination mechanism enhances the depth of understanding and memory, and fuzzy rule reasoning strengthens logical consistency and interpretability. Together, they collectively contribute to consistent performance improvements across diverse task settings.

### *D. Modal Effectiveness Analysis*

To evaluate the effectiveness of processing different modalities, we conducted modality ablation experiments on text, image, table, knowledge graph, and sequence modalities. Specifically, "Text" denotes processing using only the textual modality; "Text+Image" represents the joint processing of text and image modalities; "Text+Image+Table" refers to the collaborative processing of text, image, and table modalities; "Text+Graph" denotes the joint processing of text and graph modalities; and "Text+Graph+Sequence" represents the collaborative processing of text, graph, and sequence modalities.

First, we performed a modality effectiveness analysis on the BioMol-MQA and EHRxQA datasets. The corresponding experimental results are illustrated in Fig. 7. As shown in Fig. 7, on the MultimodalQA dataset, the F1 score indicates that the Text+Image modality improves performance by 9.43 compared with using only the Text modality, demonstrating that the image modality has a significant positive impact on model performance. Further incorporating the Table modality, the Text+Image+Table model achieves an additional improvement of 5.89. On the WebQA dataset, a similar trend is observed for the BLEU metric: adding the image modality increases the score by 19.43, and further adding the table modality results in an additional gain of 11.05.

On the BioMol-MQA dataset, the HR@5 metric shows that introducing the Graph modality representing drug–target interactions improve performance by 9.43, and further including the Sequence modality, which encodes protein amino acid sequences and drug SMILES representations, yields an additional improvement of 20.59. For the EHRxQA dataset, starting from the Image modality (chest X-ray images), incorporating the structured Table modality representing patient medical records leads to a substantial 22.00 increase in the FL metric.

These experimental results indicate that gradually introducing different modalities can significantly enhance the performance of multimodal QA models across datasets, suggesting that the model effectively mitigates modal bias during multimodal fusion. Based on these observations, the following conclusions can be drawn:

(1) Cross-modal complementarity enhances information balance. In the MultimodalQA and WebQA datasets, the text modality often dominates semantic reasoning but may overlook visual or structured information. Incorporating the Image and Table modalities provides semantic alignment and visual contextual enrichment, reducing the model's over-reliance on text and alleviating semantic bias. The consistent improvements in F1 and BLEU indicate that the model learns a more balanced feature attention distribution across the multimodal semantic space through collaborative multi-source features.

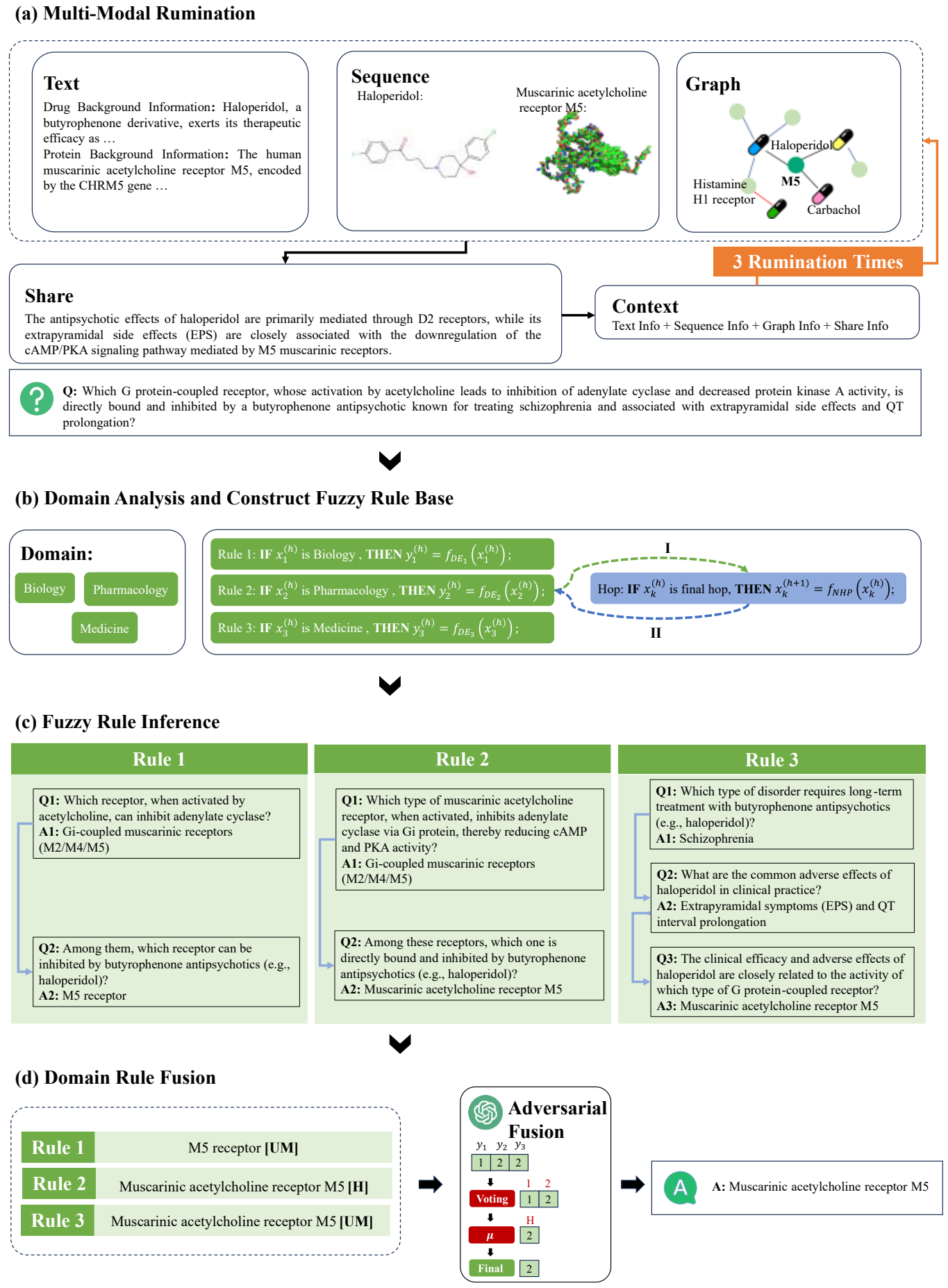

Fig. 8. Case study of the multimodal drug–target question answering process.

(2) Inter-modal mutual information improves feature discriminability. In BioMol-MQA, jointly modeling the Graph modality (drug structure graph) and the Sequence modality (protein and drug sequences) significantly improves HR@5. The graph modality encodes topological relationships between entities, while the sequence modality captures localized chemical patterns. Their fusion enhances the model's ability to distinguish multi-granularity features and aligns semantic mappings across modalities, mitigating the misleading effects of single-modality features.

(3) Structured modalities enhance model robustness and interpretability. In EHRxQA, adding the Table modality (structured medical records) improves performance by 22.00, demonstrating that structured information plays a crucial role in reducing noise from the image modality. Structured modalities provide explicit constraints and logical priors, enabling the model to reason more stably under uncertain inputs and significantly reducing generalization errors caused by modality imbalance.

(4) Modality fusion facilitates cross-modal consistency learning. Fusing features from different modalities encourages the alignment of semantic structures in a joint representation space, achieving cross-modal consistency. This process fundamentally suppresses the emergence of modal bias, allowing the model to perform integrated reasoning based on multi-source evidence rather than relying on single-modality features for shortcut learning.

In summary, multimodal fusion not only delivers substantial performance improvements but also effectively mitigates modal bias at the mechanistic level. Through cross-modal feature complementarity, semantic consistency constraints, and structured knowledge guidance, the model achieves more robust, balanced, and interpretable reasoning in complex QA tasks. These findings suggest that multimodal feature fusion is a key pathway for realizing human-like cognition and generalizable learning.

### *E. Case Study*

To provide an in-depth and intuitive illustration of MMGFS's operational mechanism and the role of each module, this section presents a case study on a QA example from BioMol-VQA, involving the drug Haloperidol and its target Muscarinic acetylcholine receptor M5. Fig. 8 presents the processing pipeline for this drug–target QA.

#### 1) Multimodal Rumination

Fig. 8 (a) illustrates the data from three modalities: text, sequence, and graph. The text modality contains background information about the drug and its corresponding target protein. The sequence modality provides the SMILES representation of the drug Haloperidol and the name of its target, Muscarinic acetylcholine receptor M5, along with the actual protein sequence obtained from UniProt. The graph modality encodes the relational edges between drug–target, drug–drug, and target–target pairs. In this case, the relations are represented as triples in the form of (subject–predicate–object), e.g., Haloperidol–binding and inhibition–Muscarinic acetylcholine receptor M5.

Each modality in Fig. 8 (a) is processed by its corresponding MIP component to extract natural semantic representations. These semantic representations are then assessed for consistency, with potential semantic conflicts identified. Once a conflict is detected, the relevant content is fed back to the MIP components of each modality to trigger a new round of semantic extraction and rumination. In this example, three rounds of rumination were performed, and the following cross-modal semantic conflicts were identified: (1) Receptor coupling conflict: D2 belongs to Gi/o coupling, while M5 belongs to Gq coupling, leading to inconsistent signaling pathways; (2) Target discrepancy: Clinically, Haloperidol primarily acts on D2 receptors, but the triple points to M5. Adverse effect mechanism conflict: EPS (extrapyramidal symptoms) is usually attributed to D2 receptors, while the text suggests involvement of M5-mediated cAMP/PKA downregulation.

After three rounds of rumination, semantic conflicts across modalities were resolved, and MMGFS extracted consistent multimodal information. This shared information, verified through cross-modal validation, possesses high confidence and is passed along with the text, sequence, and graph modality information to subsequent processing stages.

#### 2) Domain Analysis

The DA submodule identifies the core disciplines relevant to the question. Fig. 8 (b) shows that three domains were recognized: Biology, Pharmacology, and Medicine. For this question, the DA component generated three domain-specific rules and one Hop rule, all automatically created from predefined templates to form a complete GFRB (Generative Fuzzy Rule Base). This rule base not only provides structured knowledge for subsequent multimodal reasoning but also ensures that cross-domain information can be effectively integrated and inferred under rule guidance, thereby improving the accuracy and interpretability of MMGFS when handling complex questions.

#### 3) Rule-Based Reasoning

The MMFIE engine performs systematic reasoning over the domain rules during multimodal inference. Leveraging Hop rules, MMFIE iteratively generates multi-hop questions, decomposing complex problems into hierarchical sub-tasks. Meanwhile, each domain-specific rule processes its respective sub-task for fine-grained reasoning. The number of hops across domains reflects the depth and complexity of reasoning—the more hops, the longer the required knowledge integration and logical chain.

Fig. 8 (c) illustrates a concrete example: (1) Rule 1 produces the answer "M5 receptor" after two hops; (2) Rule 2 also undergoes two hops; (3) Rule 3 is processed through three hops.

The final answers from Rules 2 and 3 converge to "Muscarinic acetylcholine receptor M5". This demonstrates that multi-hop rule-based reasoning allows FAIE to integrate cross-domain knowledge effectively, maintain multimodal information consistency and accuracy, and provides interpretable, hierarchical reasoning for complex scientific questions.

#### 4) Domain Rule Fusion

DRFM integrates the reasoning results from all domain rules to produce a final high-confidence answer. The fusion process relies on a semantic voting mechanism and rule firing strength: (1) Semantic voting counts overlapping outputs from different rules, prioritizing the most frequently occurring semantic fragment; (2) Rule firing strength quantifies each rule's contribution to a specific question, allowing weighted selection of the final output.

Fig. 8 (d) provides an example where the answer "Muscarinic acetylcholine receptor M5" receives two votes from Rules 2 and 3, while "receptor M5" receives one vote from Rule 1; considering that the activation strengths of Rules 1, 2, and 3 are UM, H, and UM respectively, with H (High) exceeding UM (Upper-Medium), Rule 2 has the highest activation, making its output, "Muscarinic acetylcholine receptor M5," the final fused answer.

This fusion strategy offers multiple advantages by enhancing consistency and reliability through semantic voting and rule activation weighting, ensuring agreement among multiple rule outputs; maintaining interpretability, as each output can be traced back to specific rules and their activation weights, providing a transparent and verifiable solution for complex multimodal reasoning tasks; and mitigating hallucinations, since cross-validation and adversarial mechanisms across domain semantics help reduce spurious inferences from large language models during reasoning.

### *F. Further Tests*

Supplementary Material Part 4 provides a systematic analysis of key parameters, including the number of rules, rumination iterations, and question length, while Part 5 further evaluates the resource consumption of MMGFS, as well as its time and space complexity. The experimental and complexity analyses indicate that MMGFS maintains strong reasoning capability and modeling flexibility with controllable computational and resource costs, demonstrating good scalability and practical applicability. To verify statistical significance, we first applied the Friedman test [46] to assess overall differences among the methods, followed by the Holm post hoc test [47] for multiple comparisons. Detailed results are provided in Supplementary Material Part 6. The analysis shows that there are statistically significant differences between MMGFS and the baseline methods.

## V. Conclusion

We developed an end-to-end generative fuzzy system for MQA (MMGFS), which mitigates modal bias through collaborative multimodal processing and uses LM-based fuzzy rules with multi-hop reasoning to handle uncertainty and support deeper reasoning. Experiments on public MQA datasets show that MMGFS achieves superior performance and enhanced robustness compared with existing methods, with improved generalization and cognitive consistency.

Despite these advantages, MMGFS has several limitations. It currently supports a limited range of modalities and remains sensitive to semantic conflicts in multimodal inputs, which may affect reasoning reliability. Future work will focus on modality-agnostic representation alignment, explicit conflict-aware reasoning mechanisms, enhanced interpretability via fuzzy rule–based multi-hop inference, and adaptive continual learning to support new modalities and evolving domain knowledge.

## References

[1] H. Luo, Y. Shen, and Y. Deng, ‘Unifying text, tables, and images for multimodal question answering’, in *Findings of the Association for Computational Linguistics: EMNLP 2023*, Singapore: Association for Computational Linguistics, 2023, pp. 9355–9367.

[2] Z. Lin *et al.*, ‘Medical visual question answering: a survey’, *Artif. Intell. Med.*, vol. 143, p. 102611, Sept. 2023

[3] O. Kolomiyets and M.-F. Moens, ‘A survey on question answering technology from an information retrieval perspective’, *Inf. Sci.*, vol. 181, no. 24, pp. 5412–5434, Dec. 2011

[4] S. Antol *et al.*, ‘VQA: visual question answering’, in *Proceedings of the IEEE International Conference on Computer Vision (ICCV)*, Santiago, Chile: IEEE, Dec. 2015, pp. 2425–2433.

[5] Q. Wu, D. Teney, P. Wang, C. Shen, A. Dick, and A. Van Den Hengel, ‘Visual question answering: a survey of methods and datasets’, *Comput. Vision Image Understanding*, vol. 163, pp. 21–40, Oct. 2017

[6] S. Lu, M. Liu, L. Yin, Z. Yin, X. Liu, and W. Zheng, ‘The multi-modal fusion in visual question answering: a review of attention mechanisms’, *Peerj Comput. Sci.*, vol. 9, p. e1400, May 2023

[7] Z. Yang, X. He, J. Gao, L. Deng, and A. Smola, ‘Stacked attention networks for image question answering’, in *Proceedings of the IEEE Conference on Computer Vision and Pattern Recognition (CVPR)*, Las Vegas, NV, USA: IEEE, June 2016, pp. 21–29.

[8] A. Fukui, D. H. Park, D. Yang, A. Rohrbach, T. Darrell, and M. Rohrbach, ‘Multimodal compact bilinear pooling for visual question answering and visual grounding’, in *Proceedings of the 2016 Conference on Empirical Methods in Natural Language Processing*, Austin, Texas: Association for Computational Linguistics, 2016, pp. 457–468.

[9] Z. Yu, J. Yu, J. Fan, and D. Tao, ‘Multi-modal factorized bilinear pooling with Co-attention learning for visual question answering’, in *Proceedings of the IEEE International Conference on Computer Vision (ICCV)*, Venice: IEEE, Oct. 2017, pp. 1839–1848.

[10] A. Radford *et al.*, ‘Learning transferable visual models from natural language supervision’, in *Proceedings of the 38th International Conference on Machine Learning*, PMLR, July 2021, pp. 8748–8763.

[11] C. Jia *et al.*, ‘Scaling Up Visual and Vision-Language Representation Learning With Noisy Text Supervision’, in *Proceedings of the 38th International Conference on Machine Learning*, Virtual Only Conference: PMLR, July 2021, pp. 4904–4916.

[12] J.-B. Alayrac *et al.*, ‘Flamingo: a visual language model for few-shot learning’, in *Proceedings of the 36th International Conference on Neural Information Processing Systems*, Red Hook, NY, USA: Curran Associates Inc., Nov. 2022, pp. 23716–23736.

[13] J. Li, D. Li, S. Savarese, and S. Hoi, ‘BLIP-2: bootstrapping language-image pre-training with frozen image encoders and large language models’, in *Proceedings of the 40th International Conference on Machine Learning*, in ICML’23, vol. 202. Honolulu, Hawaii, USA: JMLR.org, July 2023, pp. 19730–19742.

[14] H. Liu, C. Li, Q. Wu, and Y. J. Lee, ‘Visual instruction tuning’, in *Proceedings of the 37th International Conference on Neural Information Processing Systems*, Red Hook, NY, USA: Curran Associates Inc., Dec. 2023, pp. 34892–34916.

[15] W. Chen, H. Hu, X. Chen, P. Verga, and W. Cohen, ‘MuRAG: multimodal retrieval-augmented generator for open question answering over images and text’, in *Proceedings of the Conference on Empirical Methods in Natural Language Processing*, Y. Goldberg, Z. Kozareva, and Y. Zhang, Eds, Abu Dhabi, United Arab Emirates: Association for Computational Linguistics, Dec. 2022, pp. 5558–5570.

[16] F. Khader *et al.*, ‘Medical diagnosis with large scale multimodal transformers: leveraging diverse data for more accurate diagnosis’, Dec. 20, 2022, *arXiv*: arXiv:2212.09162.

[17] F. Khader *et al.*, ‘Medical transformer for multimodal survival prediction in intensive care: integration of imaging and non-imaging data’, *Sci. Rep.*, vol. 13, no. 1, p. 10666, July 2023

[18] B. Yu, C. Fu, H. Yu, F. Huang, and Y. Li, ‘Unified language representation for question answering over text, tables, and images’, in *Findings of the Association for Computational Linguistics: ACL 2023*, A. Rogers, J. Boyd-Graber, and N. Okazaki, Eds, Toronto, Canada: Association for Computational Linguistics, July 2023, pp. 4756–4765.

[19] W. Liu *et al.*, ‘MMHQA-ICL: multimodal In-context learning for hybrid question answering over text, tables and images’, Sept. 09, 2023, *arXiv*: arXiv:2309.04790.

[20] Y. Wang, M. Yasunaga, H. Ren, S. Wada, and J. Leskovec, ‘VQA-GNN: reasoning with multimodal knowledge via graph neural networks for visual question answering’, in *Proceedings of the IEEE/CVF International Conference on Computer Vision (ICCV)*, Paris, France: IEEE, Oct. 2023, pp. 21525–21535.

[21] Q. Yang, Q. Chen, W. Wang, B. Hu, and M. Zhang, ‘Enhancing multi-modal multi-hop question answering via structured knowledge and unified retrieval-generation’, in *Proceedings of the 31st ACM International Conference on Multimedia*, Ottawa ON Canada: ACM, Oct. 2023, pp. 5223–5234.

[22] S. Lin, J. Hilton, and O. Evans, ‘TruthfulQA: measuring how models mimic human falsehoods’, in *Proceedings of the 60th Annual Meeting of the Association for Computational Linguistics (volume 1: Long Papers)*, S. Muresan, P. Nakov, and A. Villavicencio, Eds, Dublin, Ireland: Association for Computational Linguistics, May 2022, pp. 3214–3252.

[23] J. Wei *et al.*, ‘Chain-of-thought prompting elicits reasoning in large language models’, in *Advances in Neural Information Processing Systems*, in NIPS ’22. New Orleans, LA, USA: Curran Associates Inc., Dec. 2022, pp. 24824–24837.

[24] T. Kojima, S. (Shane) Gu, M. Reid, Y. Matsuo, and Y. Iwasawa, ‘Large language models are zero-shot reasoners’, in *Advances in Neural Information Processing Systems*, Dec. 2022, pp. 22199–22213.

[25] R. Li *et al.*, ‘AI-VQA: visual question answering based on agent interaction with interpretability’, in *Proceedings of the 30th ACM International Conference on Multimedia*, Lisboa Portugal: ACM, Oct. 2022, pp. 5274–5282.

[26] Z. Yang *et al.*, ‘MM-REACT: prompting ChatGPT for multimodal reasoning and action’, Mar. 20, 2023, *arXiv*: arXiv:2303.11381.

[27] J. Guo *et al.*, ‘From images to textual prompts: zero-shot visual question answering with frozen large language models’, in *Proceedings of the IEEE/CVF Conference on Computer Vision and Pattern Recognition (CVPR)*, Vancouver, BC, Canada: IEEE, June 2023, pp. 10867–10877.

[28] B. Jiang, Z. Zhuang, S. S. Shivakumar, D. Roth, and C. J. Taylor, 'Multi-agent VQA: exploring multi-agent foundation models in zero-shot visual question answering', Mar. 21, 2024, *arXiv*: arXiv:2403.14783.

[29] S. Singh, Y. Gupta, and S. R. Chowdhury, 'MEQA - a multi-modal interactive enterprise query answering system using multi-agent LLM', in *Proceedings of the 8th International Conference on Data Science and Management of Data (12th ACM IKDD CODS and 30th COMAD)*, Jodhpur, India: ACM, Dec. 2024, pp. 422–426.

[30] F. Wu, Z. Li, F. Wei, Y. Li, B. Ding, and J. Gao, 'Talk to right specialists: routing and planning in multi-agent system for question answering', Jan. 14, 2025, *arXiv*: arXiv:2501.07813.

[31] H. Shi *et al.*, 'MuMA-ToM: multi-modal multi-agent theory of mind', in *Proceedings of the AAAI Conference on Artificial Intelligence*, Apr. 2025, pp. 1510–1519.

[32] P. Liu *et al.*, 'HM-RAG: hierarchical multi-agent multimodal retrieval augmented generation', in *Proceedings of the 33rd ACM International Conference on Multimedia*, Dublin Ireland: ACM, Oct. 2025, pp. 2781–2790.

[33] K. S. Rajput, T. Anvekar, C. Baral, and V. Gupta, 'Rethinking information synthesis in multimodal question answering a multi-agent perspective', May 27, 2025, *arXiv*: arXiv:2505.20816.

[34] Y. Zhou, L. Song, and J. Shen, 'MAM: modular multi-agent framework for multi-modal medical diagnosis via role-specialized collaboration', in *Findings of the Association for Computational Linguistics: ACL 2025*, Vienna, Austria: Association for Computational Linguistics, July 2025, pp. 25319–25333.

[35] T. Takagi and M. Sugeno, 'Fuzzy identification of systems and its applications to modeling and control', *IEEE Trans. Syst., Man, Cybern.*, vol. SMC-15, no. 1, pp. 116–132, Jan. 1985

[36] E. H. Mamdani and S. Assilian, 'An experiment in linguistic synthesis with a fuzzy logic controller', *Int. J. Man Mach. Stud.*, vol. 7, no. 1, pp. 1–13, Jan. 1975

[37] C. Grosan and A. Abraham, 'Rule-based expert systems', *Intell. Syst.: Mod. Approach*, pp. 149–185, 2011

[38] H. Ishibuchi and T. Nakashima, 'Effect of rule weights in fuzzy rule-based classification systems', *IEEE Trans. Fuzzy Syst.*, vol. 9, no. 4, pp. 506–515, Aug. 2001

[39] Z. Zhang, L. Wen, and W. Zhao, 'Rule-KBQA: rule-guided reasoning for complex knowledge base question answering with large language models', in *Proceedings of the 31st International Conference on Computational Linguistics*, O. Rambow, L. Wanner, M. Apidianaki, H. Al-Khalifa, B. D. Eugenio, and S. Schockaert, Eds, Abu Dhabi, UAE: Association for Computational Linguistics, Jan. 2025, pp. 8399–8417.

[40] A. Mohammed, A. Sameer, and A.-R. Qasem, 'QArabPro: a rule based question answering system for reading comprehension tests in arabic', *Am. J. Appl. Sci.*, vol. 8, no. 6, pp. 652–661, June 2011

[41] N. K and M. K. M. B, 'Fuzzy rule based classifier model for evidence based clinical decision support systems', *Intell. Syst. Appl.*, vol. 22, p. 200393, June 2024

[42] A. Talmor *et al.*, 'MultimodalQA: complex question answering over text, tables and images', in *International Conference on Learning Representations*, Virtual Only Conference, May 2021.

[43] Y. Chang, M. Narang, H. Suzuki, G. Cao, J. Gao, and Y. Bisk, 'WebQA: multihop and multimodal QA', in *Proceedings of the IEEE/CVF Conference on Computer Vision and Pattern Recognition (CVPR)*, Minneapolis, MN, USA: IEEE Computer Society, Mar. 2022, pp. 16495–16504.

[44] S. Sengupta, S. Yang, P. K. Yu, F. Wang, and S. Wang, 'BioMol-MQA: a multi-modal question answering dataset for LLM reasoning over bio-molecular interactions', June 06, 2025, *arXiv*: arXiv:2506.05766.

[45] S. Bae *et al.*, 'EHRXQA: a multi-modal question answering dataset for electronic health records with chest X-ray images', in *Advances in Neural Information Processing Systems*, Red Hook, NY, USA: Curran Associates Inc., Dec. 2023, pp. 3867–3880.

[46] J. H. Friedman, 'On bias, variance, 0/1—loss, and the curse-of-dimensionality', *Data Min. Knowl. Discovery*, vol. 1, no. 1, pp. 55–77, Mar. 1997

[47] S. Holm, 'A simple sequentially rejective multiple test procedure', *Scand. J. Stat.*, vol. 6, no. 2, pp. 65–70, 1979